\documentclass[letterpaper, 10 pt, conference]{ieeeconf}  % Comment this line out if you need a4paper

\IEEEoverridecommandlockouts  
\usepackage{textcomp}

\usepackage{sansmath}
\usepackage{graphicx}
\usepackage{amsmath}
\usepackage{xspace}
\usepackage{multirow}
\usepackage{diagbox}
\usepackage{amsfonts}
\usepackage{mathtools}
\usepackage{hyperref}
\usepackage[english]{babel}
\usepackage[version=4]{mhchem}
\usepackage{siunitx}
\usepackage{blindtext}
\usepackage{hyperref}
\usepackage{longtable,tabularx}

\usepackage[utf8]{inputenc}
\usepackage{graphicx}
\usepackage{subcaption}
\usepackage[version=4]{mhchem}
\usepackage{longtable,tabularx}
\usepackage{algorithm}
\usepackage{algpseudocode}
\usepackage{float}
\usepackage{titlesec}
\usepackage{tikz}
\usepackage{pgfplots}
\pgfplotsset{compat=1.17}
\usepackage{booktabs}
\usepackage{pgfplotstable} % For handling CSV files
\usetikzlibrary{positioning, shapes.geometric, fit}
\usetikzlibrary{calc}
\usetikzlibrary{arrows.meta} 
\newcommand{\bx}{{\boldsymbol{x}}}

\newcommand{\by}{{\boldsymbol{y}}}

\newtheorem{problem}{\bf Problem}

\usepackage{dirtytalk}

\title{\LARGE \bf
Fast End-to-End Generation of Belief Space Paths for Minimum Sensing Navigation
}

\author{Lukas Taus$^{1}$, Vrushabh Zinage$^{2}$, Takashi Tanaka $^{2}$,  
Richard Tsai $^{1}$% <-this % stops a space
\thanks{$^{1}$Lukas Taus and 
Richard Tsai are with the Oden Institute for Computational Engineering and Sciences, University of Texas at Austin
        {\tt\small l.taus@utexas.edu, ytsai@math.utexas.edu}}%
\thanks{$^{2}$Vrushabh Zinage and Takashi Tanaka are with the Department of Aerospace Engineering and Engineering Mechanics, University of Texas at Austin
        {\tt\small vrushabh.zinage@utexas.edu, ttanaka@utexas.edu}}%
}

\begin{document}

\bibliographystyle{IEEEtran}

\maketitle
\thispagestyle{empty}
\pagestyle{empty}

%%%%%%%%%%%%%%%%%%%%%%%%%%%%%%%%%%%%%%%%%%%%%%%%%%%%%%%%%%%%%%%%%%%%%%%%%%%%%%%%
\begin{abstract}
We revisit the problem of motion planning in the Gaussian belief space. Motivated by the fact that most existing sampling-based planners suffer from high computational costs due to the high-dimensional nature of the problem, we propose an approach that leverages a deep learning model to predict optimal path candidates directly from the problem description. Our proposed approach consists of three steps. First, we prepare a training dataset comprising a large number of input-output pairs: the input image encodes the problem to be solved (e.g., start states, goal states, and obstacle locations), whereas the output image encodes the solution (i.e., the ground truth of the shortest path). Any existing planner can be used to generate this training dataset. Next, we leverage the U-Net architecture to learn the dependencies between the input and output data. Finally, a trained U-Net model is applied to a new problem encoded as an input image. 
From the U-Net's output image, which is interpreted as a distribution of 
paths,
%to predict the corresponding output image, from which 
an optimal path candidate is reconstructed. 
%Numerical simulations suggest that 
The proposed method significantly reduces computation time compared to the sampling-based baseline algorithm.

% In this paper, we consider the problem of motion planning in the presence of Gaussian uncertainty i.e. belief space planning. Motivated by the fact that most conventional methods that generate belief space paths are computationally expensive or focus on improving the sampling efficiency of these planners, we particularly explore the feasibility of directly predicting the optimal feasible path, given the start and the goal configurations, and the obstacle information.  Our proposed approach mainly consists of three steps. In the first step, we collect the input-output data from traditional belief space planners where the input data consists of the start states, goal states, and the obstacles information. The output data consists of the corresponding information about the optimal feasible paths that are obtained via an a-priori known sampling-based planner. In the second step, we leverage the U-Net architecture to learn the dependencies of the input data to the output data. Finally, we propose an algorithm for reconstruction of the optimal path from the image space to the configuration space. Numerical simulations are presented to verify the efficacy of the proposed approach.

\end{abstract}

%%%%%%%%%%%%%%%%%%%%%%%%%%%%%%%%%%%%%%%%%%%%%%%%%%%%%%%%%%%%%%%%%%%%%%%%%%%%%%%%
\section{Introduction\label{sec:introduction}}
We consider the problem of improving the computational efficiency of belief space planning algorithms. Belief space planning is crucial for robotic systems operating in uncertain environments, with applications ranging from autonomous driving \cite{lozano2012autonomous_general_2} to various space applications \cite{volpe2003rover_general_3}. However, the computational complexity of belief space planning, particularly in the presence of Gaussian uncertainty \cite{agha2014firm,zheng2024cs_brm}, remains a significant challenge for real-time implementation in many robotic applications. We address this challenge by exploring a novel approach to belief space planning that leverages Deep Learning techniques to directly predict optimal feasible paths under Gaussian uncertainty. While traditional methods for generating belief space paths are often computationally expensive,  our approach constructs optimal paths directly from the density of feasible paths inferred by a trained artificial neural network from the problem inputs.

Motion planning determines a path/trajectory that avoids collisions, from a starting configuration to a target configuration \cite{lavalle2006planning_general_1}. 
To meet the demand for efficient real-time algorithms, various sampling-based motion planning (SMP) methods have been developed, including Rapidly-exploring Random Trees (RRT) \cite{lavalle1998rapidly_rrt}, RRT* \cite{gammell2014informed_rrt_star}, Potentially guided-RRT* (P-RRT*) \cite{qureshi2016potential_prrt_star}, and their bi-directional variants~\cite{tahir2018potentially_prrt_star_2}. Additional approaches like BIT*~\cite{gammell2020batch_bit_star}, Advanced BIT* \cite{strub2020advanced_abit_star}, and RABIT*~\cite{choudhury2016regionally_rabit_star} have also been proposed. However, despite these advancements, these algorithms still face challenges in scaling to high-dimensional spaces often encountered in real-world applications and remain computationally expensive.

One severe limitation of SMPs is the increase in the number of samples required as the dimensionality of the configuration space (C-space) grows \cite{hsu1997path_limitation_1}. Several works have employed deep learning to prioritize samples in C-space to address this. For instance, \cite{ichter2018learning_bias_1} utilized the latent space from a conditional variational autoencoder (CVAE) to steer sampling towards essential samples. RL-RRT \cite{chiang2019rl_bias_2} applies deep reinforcement learning (RL) to direct the expansion of the tree towards advantageous areas of the C-space. Furthermore, \cite{zhang2018learning_bias_3} adopts RL to create an implicit sampling distribution that lowers the number of samples needed.

Recent works have also addressed the computational bottleneck arising from collision-checking modules. \cite{das2020learning_collision_1} and \cite{kew2019neural_collsion_2} use learned function approximators to improve collision detection modules for SMPs. Value Iteration Networks \cite{tamar2016value_collision_3} use convolutional neural networks (CNNs) to process discrete maps and predict movement policies. Generalized Value Iteration Networks \cite{niu2018generalized_collision_4} expand upon this by employing graph neural networks (GNNs) for more complex graph structures. Our work is complementary to these approaches. Instead of improving the sampling efficiency or the computational complexity of SMPs, we explore the possibility of directly predicting the optimal feasible path given the start states, goal states, and obstacle information in belief space. Towards this goal, the contributions of this paper include:
\begin{enumerate}
    \item We propose a novel deep learning-based framework for belief space planning that leverages a U-Net architecture to directly predict optimal feasible paths under Gaussian uncertainty. This approach significantly reduces computational complexity compared to traditional sampling-based methods in belief space.
    \item We develop an algorithm for reconstructing optimal paths from the image space to the configuration space, bridging the gap between the neural network's output and the practical application of the predicted paths.
\end{enumerate}
The overall structure of the paper is as follows. Section \ref{sec:problem_statement} discusses the problem statement followed by the proposed approach in Section \ref{sec:proposed_approach}. Section \ref{sec:results} discusses the results followed by concluding remarks in Section \ref{sec:conclusion}.
% \section{Preliminaries\label{sec:prelim}}
\section{Problem Statement\label{sec:problem_statement}}

{While the proposed approach applies to a broader class of motion planning scenarios, our study in this paper specifically focuses on the Gaussian belief space planning for minimum sensing navigation \cite{pedram2021rationally, pedram2022gaussian_ali_tro}. In \cite{pedram2022gaussian_ali_tro}, the authors formulated the planning problem for a traveling agent subject to Gaussian disturbances as a path planning problem over the space of belief states $b:=\{\bx,\;P\}$, where $\bx\in\mathcal{X}\subset\mathbb{R}^d$ is the state and $P\in \mathbb{S}^d_{++}$ is the associated covariance matrix. Denote by $\gamma(t)=(\bx(t), P(t)), 0\leq t \leq T$, the path to be designed in the Gaussian belief space $\mathbb{B} = \mathbb{R}^d \times \mathbb{S}^d_{++}$ 
such that $\gamma(0)=b_0$ (given initial configuration) and $\gamma(T) \in \mathbb{B}_{\text{target}}$ (given set of target configurations). This design aims to minimize the path cost $c(\gamma)$.
In \cite{pedram2022gaussian_ali_tro}, $c(\gamma)$ is chosen to minimize the path length of $\gamma$ and the expected sensing cost quantified by the end-to-end information gain:
\begin{equation}
\label{eq:path_length}
c(\gamma)\!=\!
\left(\!\begin{array}{c} 
%\text{Euclidean length} \\ 
\text{Path length of $\gamma$}  \end{array}\!\right)\!+\!\alpha \!
\left(\!\begin{array}{c} \text{Information gain} \\ \text{needed to follow} \\ \text{the path plan $\gamma$} \end{array}\!\right),
\end{equation}
where $\alpha\geq 0$ is a user-defined weight factor. In this context, the path length is evaluated solely within the state space $\mathcal{X}$, while the information gain is measured exclusively in the space of covariance matrices $\mathbb{S}^d_{++}$. Without providing the details of \eqref{eq:path_length} (which can be found in \cite{pedram2022gaussian_ali_tro}), we formulate the problem considered as following:
\begin{problem}\label{prob}
\normalfont
Given an initial belief state $ b_0 = \left(\bx_0, P_0\right) \in \mathbb{B}$, a closed subset $ \mathbb{B}_{\text{target}} \subset \mathbb{B} $ denoting the target belief region, and a set of obstacles $ \mathcal{X}_{\text{obs}}^j \subset \mathbb{R}^d $ for $ j \in [1;M]_d$, find the optimal path  $\gamma^\star$ that is collision-free with a confidence level $ \chi^2 > 0 $:
\begin{align}
\begin{array}{cl}
\gamma^\star=& \underset{\gamma}{\text{argmin}} \;\; c(\gamma) \\
& \text{s.t. } \gamma(0) = b_0,\nonumber\\
&\;\;\quad \gamma(T) \in \mathbb{B}_{\text{target}} \\
& \;\;\quad\left(\bx(t) - \bx_{\text{obs}}\right)^{\top} P^{-1}(t) \left(\bx(t) - \bx_{\text{obs}}\right) \geq \chi^2 \\
& \;\;\quad\forall \;t \in [0, T], \quad \forall \bx_{\text{obs}} \in \mathcal{X}_{\text{obs}}^j, \quad \forall j \in [1;M]_d.
\end{array}
\label{eqn:problem_statement}
\end{align}
\label{prob:problem_statement}
\end{problem}
}

\begin{figure}
    \centering
    \includegraphics[width=\linewidth]{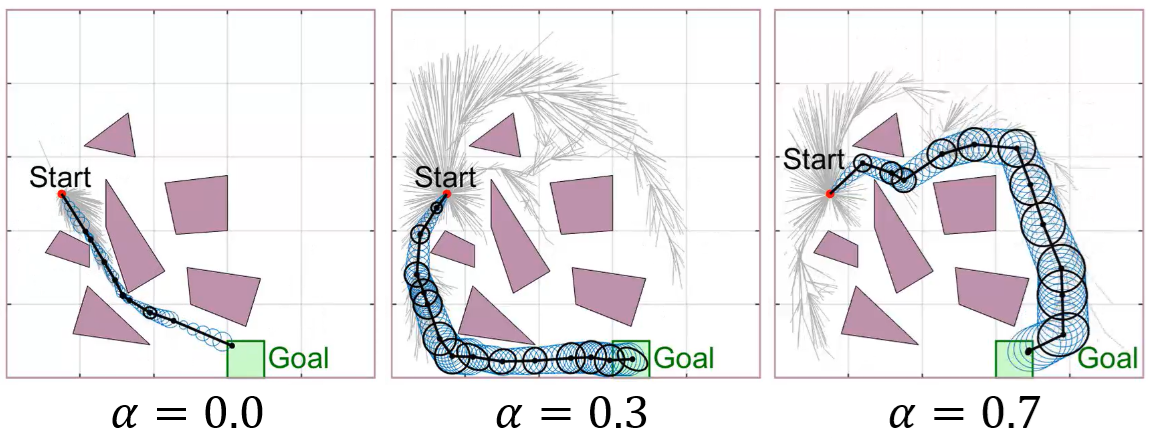}
    \caption{The RI-RRT* algorithm applied to Problem 1.}
    \label{fig:rirrt}
\end{figure}

Figure~\ref{fig:rirrt} shows the results of solving Problem 1 under different values of $\alpha$.
When $\alpha=0$, $\gamma^*$ is the shortest path in the standard sense. When $\alpha>0$ is large, $\gamma^*$ tries to minimize the information gain by maintaining a large uncertainty ($\propto \log\det P(t)$). Consequently, $\gamma^*$ avoids narrow passages.

{The work
\cite{pedram2021rationally,pedram2022gaussian_ali_tro} applies a variant of the RRT* algorithm (termed the Rationally Inattentive RRT*, or RI-RRT* algorithm) to numerically solve Problem 1. Unfortunately, it is difficult to run RI-RRT* within the practical time scale in many applications due to the high dimensional nature of Problem 1. Notice that if $d=2$, then $\mathbb{B}$ is a five-dimensional space.
We propose an approach that bypasses RI-RRT* by leveraging a U-Net model to predict $\gamma^*$.
}

\section{The Proposed Approach\label{sec:proposed_approach}}
Our proposed approach for belief space planning consists of three main steps: dataset collection, path prediction using a deep learning model, and reconstruction of the path from the image to the configuration space. In the dataset collection phase, we utilize a traditional belief space planner such as RI-RRT* to generate a dataset. This dataset comprises input data (start configurations, goal configurations, and obstacle information represented as occupancy grids) and corresponding output data (optimal feasible paths generated by the planner). 
Specifically, we generate images with various start, goal, and obstacle configurations, and run the RI-RRT* planner for each scenario to obtain the optimal feasible path under Gaussian uncertainty in the configuration space.
% Particularly, we generate multiple images consisting of various start and goal configurations, as well as different obstacle configurations, initiating the traditional planner RI-RRT* for each scenario to obtain the optimal feasible path that assumes Gaussian uncertainty in the configuration space.

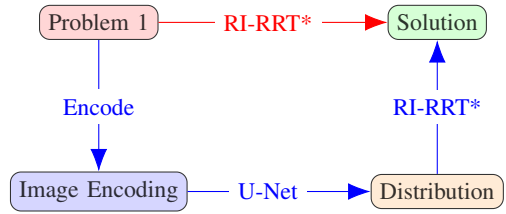
\begin{figure}
    \centering
\begin{tikzpicture}[scale=0.75]
\tikzstyle{every node}=[font=\small]
        % Define nodes with curved corners and different opaque colors
        \node[draw, rounded corners, fill=red!20, opacity=0.8] (P) at (0,0) {Problem \ref{prob}};
        \node[draw, rounded corners, fill=green!20, opacity=0.8] (S) at (6,0) {Solution};
        \node[draw, rounded corners, fill=blue!20, opacity=0.8] (E) at (0,-3) {Image Encoding};
        \node[draw, rounded corners, fill=orange!20, opacity=0.8] (D) at (6,-3) {Distribution};

        % Draw arrows with text
        \draw[-{Latex[length=3mm]}, color=red] (P) -- (S);
        \node[fill=white, text=red] at (3,0) {RI-RRT*};
        \draw[-{Latex[length=3mm]}, color=blue] (P) -- (E);
        \node[fill=white, text=blue] at (0,-1.5) {Encode};
        \draw[-{Latex[length=3mm]}, color=blue] (E) -- (D);
        \node[fill=white, text=blue] at (3,-3) {U-Net};
        \draw[-{Latex[length=3mm]}, color=blue] (D) -- (S);
        \node[fill=white, text=blue] at (6,-1.5) {RI-RRT*};
\end{tikzpicture}
    \caption{Our approach replaces the traditional algorithm (RI-RRT*) shown in red with the three steps in blue.}
    \label{fig:approach}
\end{figure}

For the path prediction task, we employ a neural network using the U-Net architecture, which has shown promising results in image-to-image translation problems \cite{unet}. The model takes as input a multi-channel image containing the obstacle map, start configuration, and goal configuration. It outputs a single-channel image representing the predicted optimal path. The U-Net's encoder-decoder structure allows the model to capture both low-level and high-level features of the input data. We train the model using the collected dataset, minimizing the binary entropy loss between the predicted paths and the ground truth paths generated by the traditional planner. During inference, given a new scenario, the trained model can quickly predict the optimal feasible path without the need for computationally expensive sampling or search procedures, significantly reducing the time required for belief space planning while maintaining the same or similar path quality. After the U-Net model predicts the optimal feasible path in the image space, the crucial final step is 
to reconstruct the path 
%in
%the reconstruction of this path 
%in the configuration space. The reconstruction process translates the predicted path from a 2D image representation back 
into a sequence of configurations a robot can execute (see Fig.~\ref{fig:approach}).

% Our proposed approach consists of replacing traditional belief space planners such as RI-RRT* with a 3 step approach. First we encode the available information on start configuration, goal configuration and obstacle information into images using occupancy grids and distance functions. This encoding can then be used as an input of a neural network that has been trained on predicting optimal paths from generated data in form of a probability distribution. This probability distribution can then be used to draw sample points in RI-RRT* algorithm in a more informed way which yields more accurate results with less computational complexity since less sample points are needed. We illustrate this approach in Fig. \ref{fig:approach}.

%\subsection{Learning-based Framework for Path Prediction and Reconstruction}
%\subsection{Learning-based Framework}
Now we discuss the three steps outlined above. %in detail.
\subsection{Data Collection}
To generate training data for our neural network, we utilize the Rationally Inattentive RRT* (RI-RRT*) algorithm \cite{pedram2022gaussian_ali_tro}, a sampling-based planner for belief space. RI-RRT* extends the RRT* algorithm to account for Gaussian uncertainty in the robot's state. For each planning scenario, we sample from the obstacle configuration, and for each sample we generate multiple instances with a fixed goal and various start configurations. We then execute RI-RRT* for each instance to obtain the optimal feasible path in the belief space. The input data consists of the start configuration, goal configuration, and obstacle information represented as occupancy grids. The corresponding output data comprises the optimal feasible paths generated by RI-RRT*. This dataset serves as an input for training our U-Net architecture to predict optimal paths directly from problem inputs.
\subsection{Path Prediction}
\textit{Image Encoding:}
We encode the available information into images utilizing occupancy grids with resolution $(N_1, N_2)$. This approach discretizes the state space $[0,1]^2$ on the grid:
%uniform Cartesian grid:
\begin{equation}
    \left\{\left(\frac{i}{N_1-1}, \frac{j}{N_2-1}\right)\right\}_{\substack{i \in \{0,\dots,N_1-1\},\\j \in \{0,\dots,N_2-1\}}}\nonumber
\end{equation}
The obstacle encoding employs an indicator function on the positions of occlusions in the state space:
\begin{equation}
    O_{i,j} = \begin{cases} 
        1 & \text{if } \begin{bmatrix} \frac{i}{N_1-1}\\ \frac{j}{N_2 - 1}\end{bmatrix} \in \bigcup_{k \in [1;M]_d} \mathcal{X}_{\text{obs}}^k\\ 
        0 & \text{else}
    \end{cases}
\end{equation}
For the initial state and target area, we utilize a distance function encoding to provide richer information for neural network processing. The target area encoding is defined as:
\begin{equation}
    T_{i,j} = \inf_{y \in \mathbb{B}_{\text{target}}} \left\Vert \begin{bmatrix}
        \frac{i}{N_1-1} \\ \frac{j}{N_2-1}
    \end{bmatrix} - y \right\Vert_2
\end{equation}
The initial position encoding is given by:
\begin{equation}
    I_{i,j} = \left\Vert \begin{bmatrix}
        \frac{i}{N_1-1} \\ \frac{j}{N_2-1}
    \end{bmatrix} - \bx_0 \right\Vert_2
\end{equation}

\begin{figure}
    \centering
    \includegraphics[width=\linewidth]{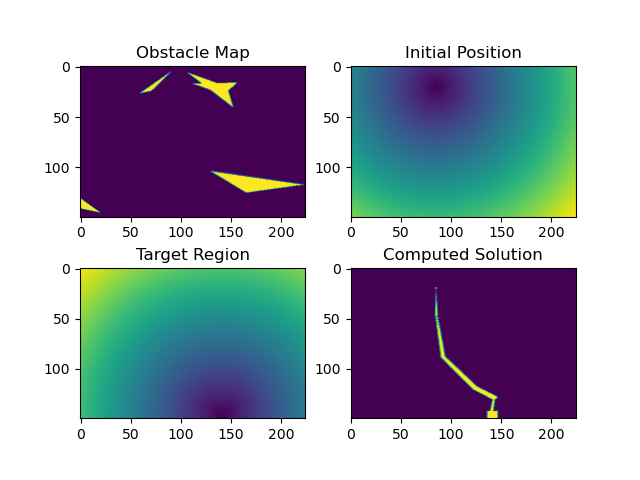}
    \caption{Visual representation of the encoding scheme for available information and the computed optimal path}
    \label{fig:encoding}
\end{figure}
Fig.~\ref{fig:encoding} illustrates 
three input images encoding the problem, as well as the label image depicting the optimal path computed by RI-RRT*. Since it is natural to represent the optimal path $\gamma^\star(t) = (\bx_t, P(t))$, $t \in [0,T]$ as a ``tube" with varying widths (the trace of the confidence ellipses shown in Fig.\ref{fig:rirrt}), we defined the label image to encode the geometry of $\gamma^\star$ using an indicator function:
\begin{equation}
    L_{i,j} = \begin{cases}
        1 &\text{if } d_t\left(\begin{bmatrix}\frac{i}{N_1-1}\\\frac{j}{N_2-1}\end{bmatrix}, \gamma^* \right) < \chi^2 \text{ for some } t \in [0,T]\\
        0 & \text{otherwise}
    \end{cases}
\end{equation}
where $d_t(\bx , \gamma) = (\bx(t) - \bx)^\top P(t)^{-1} (\bx(t) - \bx)$.
This encoding allows for the interpretation of approximating a probability distribution, as the probability of a pixel (or grid point) being part of the path is $1$ if it is within the ``tube" region and $0$ otherwise, given the optimal path.

\begin{figure}[ht!]
    \centering
    \begin{tikzpicture}
    [scale = 0.65]
    \tikzstyle{every node}=[font=\small]
    % First block with curved corners and opacity
    \filldraw[color=black!60, fill=orange!30, opacity=0.7, ultra thick, rounded corners=5pt] (-0.3, 0) rectangle (2.3, 1);
    \draw[->, color=red!60, ultra thick] (1, 0) -- (1, -1.99);
    \node at (1,0.5) {Input};
    
    \filldraw[color=red!60, fill=red!5, opacity=0.7, ultra thick] (1,-1) circle (0.5);
    \node at (1,-1) {Down};
    
    \filldraw[color=black!60, fill=black!20, opacity=0.7, ultra thick, rounded corners=5pt] (-0.3, -3) rectangle (2.3, -2);
    \node at (1,-2.5) {Encode 1};
    
    \draw[->, color=red!60, ultra thick] (1, -3) -- (1, -4.99);
    \filldraw[color=red!60, fill=red!5, opacity=0.7, ultra thick] (1,-4) circle (0.5);
    \node at (1,-4) {Down};
    
    \filldraw[color=black!60, fill=black!20, opacity=0.7, ultra thick, rounded corners=5pt] (-0.3, -6) rectangle (2.3, -5);
    \node at (1,-5.5) {Encode 2};
    
    \draw[->, dashed, color=red!60, ultra thick] (1, -6) -- (1, -7.99);
    
    \filldraw[color=black!60, fill=black!20, opacity=0.7, ultra thick, rounded corners=5pt] (-0.3, -9) rectangle (2.3, -8);
    \node at (1,-8.5) {Encode D-1};
    
    \draw[->, color=red!60, ultra thick] (1, -9) -- (1, -10.99);
    \filldraw[color=red!60, fill=red!5, opacity=0.7, ultra thick] (1,-10) circle (0.5);
    \node at (1,-10) {Down};
    
    \filldraw[color=black!60, fill=black!20, opacity=0.7, ultra thick, rounded corners=5pt] (-0.3, -12) rectangle (2.3, -11);
    \node at (1,-11.5) {Encode D};
    
    % Horizontal arrow and bottleneck
    \draw[->, color=violet!60, ultra thick] (2.3, -11.5) -- (5, -11.5) -- (5, -10.5);
    \filldraw[color=violet!60, fill=violet!5, opacity=0.7, ultra thick] (3.5,-11.5) circle (0.5);
    \node at (3.5,-11.5) {BN};
    
    \filldraw[color=black!60, fill=black!20, opacity=0.7, ultra thick, rounded corners=5pt] (3.7, -10.5) rectangle (6.3, -9.5);
    \node at (5,-10) {Decode D};
    
    \draw[->, color=blue!60, ultra thick] (5, -9.5) -- (5, -7.5);
    \draw[color=green!60, ultra thick] (2.3, -8.5) -- (4.5, -8.5);
    
    \filldraw[color=blue!60, fill=blue!5, opacity=0.7, ultra thick] (5,-8.5) circle (0.5);
    \node at (5,-8.5) {Up};
    
    \filldraw[color=black!60, fill=black!20, opacity=0.7, ultra thick, rounded corners=5pt] (3.7, -7.5) rectangle (6.3, -6.5);
    \node at (5,-7) {\small Decode D-1};
    
    \draw[->, dashed, color=blue!60, ultra thick] (5, -6.5) -- (5, -4.5);
    \draw[dashed, color=green!60, ultra thick] (2.3, -5.5) -- (5, -5.5);
    
    \filldraw[color=black!60, fill=black!20, opacity=0.7, ultra thick, rounded corners=5pt] (3.7, -4.5) rectangle (6.3, -3.5);
    \node at (5,-4) {Decode 2};
    
    \draw[->, color=blue!60, ultra thick]  (5, -3.5) -- (5, -1.5);
    \draw[color=green!60, ultra thick] (2.3, -2.5) -- (4.5, -2.5);
    
    \filldraw[color=blue!60, fill=blue!5, opacity=0.7, ultra thick] (5,-2.5) circle (0.5);
    \node at (5,-2.5) {Up};
    
    \filldraw[color=black!60, fill=black!20, opacity=0.7, ultra thick, rounded corners=5pt] (3.7, -0.5) rectangle (6.3, -1.5);
    \node at (5,-1) {Decode 1};
    
    \draw[->, color=blue!60, ultra thick] (5, -0.5) -- (5, 0.5) -- (6.7, 0.5);
    \draw[color=green!60, ultra thick] (2.3, 0.5) -- (4.5, 0.5);
    
    \filldraw[color=blue!60, fill=blue!5, opacity=0.7, ultra thick] (5,0.5) circle (0.5);
    \node at (5,0.5) {Up};
    
    \filldraw[color=black!60, fill=black!20, opacity=0.7, ultra thick, rounded corners=5pt] (6.7, 0) rectangle (9.3, 1);
    \node at (8,0.5) {Decode 0};
    
    \draw[->, color=orange!60, ultra thick] (8, 0) -- (8, -2);
    \filldraw[color=orange!60, fill=orange!5, opacity=0.7, ultra thick] (8,-1) circle (0.5);
    \node at (8,-1) {FC};
    
    \filldraw[color=black!60, fill=orange!30, opacity=0.7, ultra thick, rounded corners=5pt] (6.7, -3) rectangle (9.3, -2);
    \node at (8, -2.5) {Output};
    
    % Legend
    \draw[color=black!60, ultra thick, rounded corners=5pt] (6.5, -12) rectangle (9.5, -9);
    \draw[color=green!60, ultra thick] (6.75, -11.5) -- (7.25, -11.5);
    \node at (8.25, -11.5) {skip con.};
    
    \draw[color=violet!60, ultra thick] (6.75, -11) -- (7.25, -11);
    \node at (8.35, -10.99) {bottleneck};
    
    \draw[color=orange!60, ultra thick] (6.75, -10.5) -- (7.25, -10.5);
    \node at (8.35, -10.5) {fin. conv.};
    
    \draw[color=blue!60, ultra thick] (6.75, -10) -- (7.25, -10);
    \node at (8.35, -10.02) {tran. up};
    
    \draw[color=red!60, ultra thick] (6.75, -9.5) -- (7.25, -9.5);
    \node at (8.35, -9.49) {tran. down};
\end{tikzpicture}
    \caption{Full architecture of a UNet. The diagram illustrates the sequence of encoding and decoding blocks, skip connections, and the bottleneck layer. Each block represents a distinct operation, with arrows indicating data flow.}
    \label{fig:unet_arch}
\end{figure}
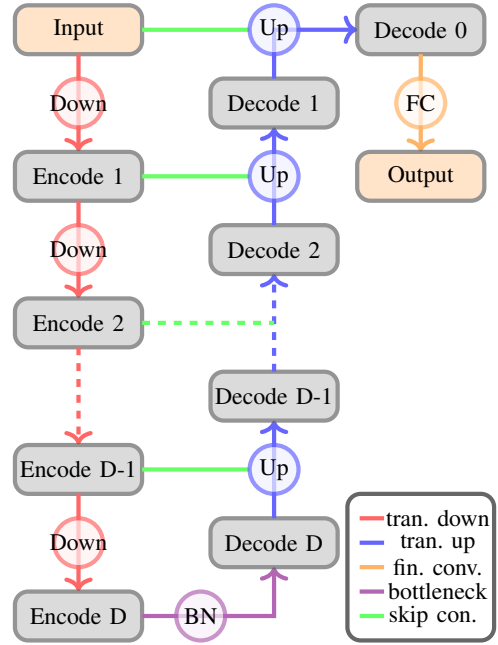

\textit{Neural Network Approximation \label{sec:NN_description}:}
% \label{sec:NN_description}
For our neural network architecture, we employ the U-Net \cite{unet}, which has demonstrated remarkable success in image processing tasks. The fundamental structure of the U-Net architecture is illustrated in Fig. \ref{fig:unet_arch}.
The transition-down block effectively separates the highest frequency components of its input, storing this information in separate channels. The output channel count doubles, which can be interpreted as splitting the high-frequency components of each channel and storing them separately. However, the convolution layer is more versatile, capable of separating high-frequency components and simultaneously applying transformations to these and the remaining channels. The max pooling layer serves to extract crucial information while discarding data irrelevant to optimal path approximation.

Conversely, the transition-up block reconstructs the information separated by the transition-down block into an image with fewer channels. This process can be viewed as merging information from two channels into one, with each reconstructed channel having access to information from all other channels, consolidating the information derived from the filters in the U-Net.

\begin{figure} %[H]
    \centering
    \includegraphics[width=\linewidth]{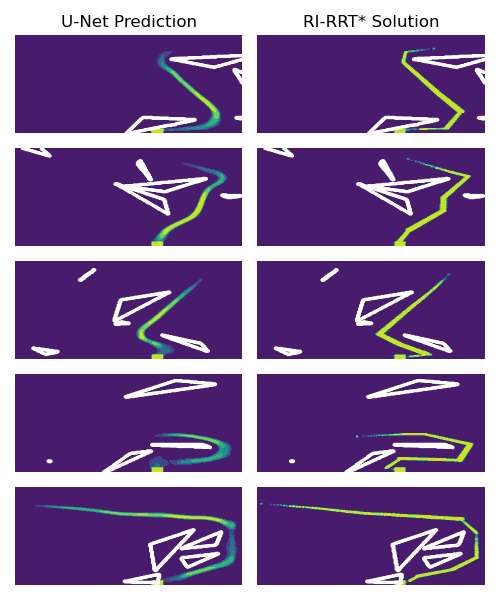}
    \caption{The left column shows the probability density predicted by the neural network and the right column shows the reference solution computed by RI-RRT* for the same obstacle map, start, and target locations.}
    \label{fig:unet_test_data}
\end{figure}

The bottleneck layer, applied when frequency components are most separated, can be considered an additional transformation on all different frequencies. The final convolution acts as a terminal transformation, pooling all available transformed information into the final image. The network architecture begins with the application of multiple transition-down blocks to separate different frequency components of the available information. This process is repeated D (``Depth") times until the input is transformed into ``Encode D". At this point, the bottleneck layer processes the separated information before reconstruction by the transition-up layers. The skip connections (denoted by green lines) allow transition-up blocks to access previous information, addressing the potential loss of crucial reconstruction data due to max pooling in the transition-down blocks. Finally, the terminal convolution is applied to further refine the result. The network $f_\theta$ is trained using a variant of the gradient descent algorithm, optimizing the following problem $\min_\theta \;\mathrm{BCE}(f_\theta(O, T, I), L)$
% \begin{align}
% \min_\theta \;\mathrm{BCE}(f_\theta(O, T, I), L)
% \label{eq:net_training}
% \end{align}
where $\mathrm{BCE}$ represents the Binary Cross-Entropy loss:
\begin{align}
\mathrm{BCE}(\bx, \by) = \frac{1}{L}\sum_{i=1}^L y_i\log(x_i) + (1-y_i)\log(1-x_i)
\end{align}
for $\bx, \by \in \mathbb{R}^L$.
We chose this specific loss function to mitigate vanishing gradient issues. Given that the label matrix $L$ has values between 0 and 1, the sigmoid function is a natural choice for the terminal activation function of the neural network. However, the sigmoid function is extremely flat away from 0, resulting in gradients close to 0. The BCE loss function effectively addresses this issue by essentially inverting the sigmoid function before measuring the distance to the label.
Upon successful training, the network can accurately predict probability distributions for the optimal path regions, as illustrated in Fig. \ref{fig:unet_test_data}.
\subsection{Path Reconstruction Algorithm}
\begin{figure}%[H]
    \centering
    \includegraphics[width=\linewidth]{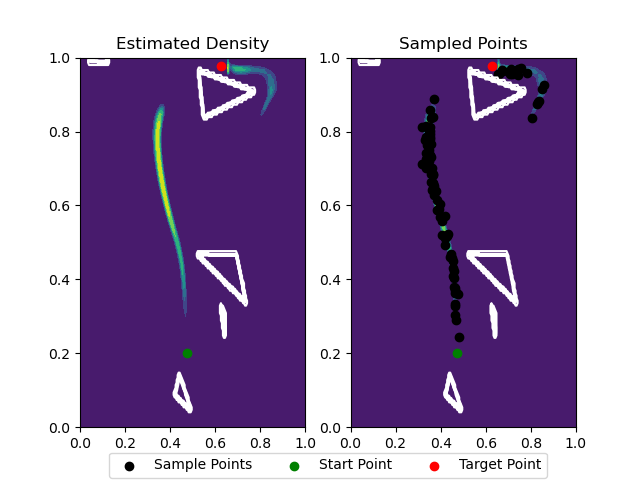}
    \caption{$100$ sample points drawn from the path distribution predicted by the neural network.}
    \label{fig:samplepoints}
\end{figure}

We leverage the predicted path from the neural network to enhance the efficiency of the RI-RRT* algorithm in computing near-optimal paths. The network's output is interpreted as a probability density function over $[0,1]^2$, from which we draw random sample points, as illustrated in Fig. \ref{fig:samplepoints}. However, these sample points do not directly provide information about the required covariance matrices.

\begin{figure}%[H]
    \centering
    \includegraphics[width=\linewidth]{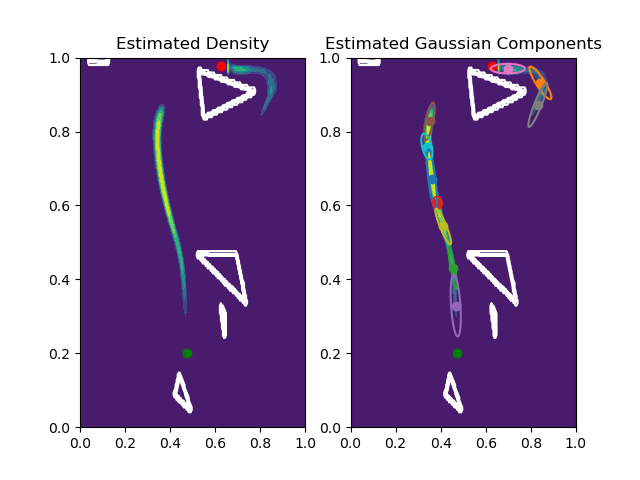}
    \caption{Estimated Gaussian components with covariance matrices of the sample points. Covariance matrices, illustrated by the ellipses on the right subplot, capture the thickness of the predicted path. }
    \label{fig:gaussian_comp}
\end{figure}

\begin{figure}%[H]
    \centering
    \includegraphics[width=\linewidth]{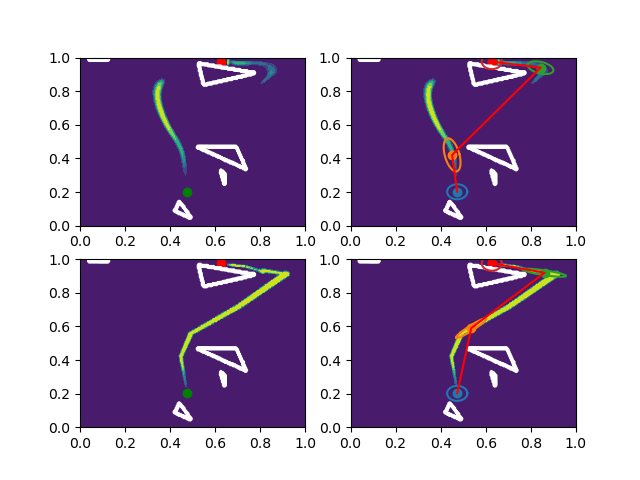}
    \caption{Reconstructed path using the proposed algorithm}
    \label{fig:path_recon}
\end{figure}

Through this we obtain a more informative initial set of sample points $(x_i, P_i)$ which allows sample-based optimization algorithms (like RI-RRT*) to operate with significantly fewer samples, leading to a significant speed up. Using these pairs we can construct a graph with $(x_i, P_i)$ as vertices and edge values related to the objective function of problem \ref{prob}. After applying an optimization algorithm to the graph, we obtain an estimate for a near-optimal path. An example of a reconstructed path using this method is presented in Fig. \ref{fig:path_recon}.
We employ a Gaussian mixture model that uses the sampled points to compute pairs of means and covariance matrices (10 in our analysis). This strategy is reasonable since the thickness of the path is related to the covariance of the optimal path in this location by the encoding of the training data. An example of the resulting Gaussian components is depicted in Fig.~\ref{fig:gaussian_comp}.

It is important to note that during graph construction, collision checks are performed, and edges intersecting obstacles are removed, which may lead to a disconnected graph. In cases where the start and target nodes are not connected, the path reconstruction fails. However our numerical experiments in Section \ref{sec:results} demonstrate a 96\% success rate when using the network prediction as a base distribution. In the rare cases where reconstruction fails, additional nodes can be sampled (as done in RI-RRT*) until the completeness of the graph is ensured. 


\section{Results\label{sec:results}}
This section presents numerical experiments comparing the accuracy and computational complexity of our proposed approach with the RI-RRT* algorithm. We also examine the feasibility of our approach when encountering scenarios outside the neural network's training data.
\subsection{Experiment Setup}
Our experimental setup consists of an environment with $5$ randomly generated triangular obstacles. The target area is fixed, while the initial position is randomly sampled. Fig. \ref{fig:encoding} illustrates an example of this environment, including the initial state and target, using our encoding scheme. We generated 19,684 such examples, along with their solutions using RI-RRT* algorithm \cite{pedram2022gaussian_ali_tro}, to train the neural network. An additional 100 examples were used to analyze the network's performance in terms of accuracy, generalizability, and computational cost.

\begin{figure}
    \centering
    \includegraphics[width=\linewidth]{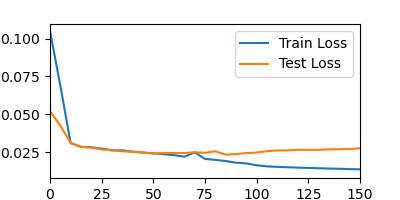}
    \caption{Values of the training and testing losses. The x-axis depicts the number of training epochs.}
    \label{fig:training}
\end{figure}

As described in Section \ref{sec:NN_description}, we employ a U-Net architecture to approximate the distribution of the optimal path region. For details on the architecture, we refer the reader to the Python code available at \cite{Code}. The neural network was trained on the $19,684$ generated examples using the Adam optimizer \cite{kingma2017adammethodstochasticoptimization} with a batch size of $32$. Throughout the training process, we monitored the Binary Cross-Entropy (BCE) loss function for both the training examples and an additional 100 different examples (test data) generated in the same manner. Fig. \ref{fig:training} illustrates the evolution of the loss function for these two datasets as the optimization algorithm progresses.
In Fig. \ref{fig:training} the optimization algorithm is applied with a learning rate of $10^{-4}$ for the first 100 epochs and $10^{-5}$ afterward. 
We observed that the neural network begins to overfit after 95 epochs, so we chose to use the weights obtained at that point.\\

\subsection{Generalizability}
In Fig. \ref{fig:unet_test_data}, we observed examples of predicted paths, which demonstrated the network's capability to produce reasonable-looking results. However, these results were generated from obstacle geometries similar to those used in training. To analyze the effects of deviating obstacle geometry on the neural network prediction, we introduce an additional circle-shaped obstacle, which is randomly placed in the obstacle map. The geometric properties of this modified obstacle map differ significantly from those in the original training data set.

\begin{figure}[]
    \centering
    \includegraphics[width=\linewidth]{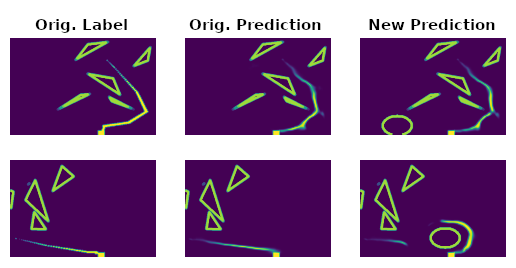}
    \caption{Predicted path distributions with the presence of a random circle to the obstacle map. The first row shows that the network prediction is stable when the circle does not obstruct the original path distribution. The second row shows the capability of the network to re-route if necessary. Note that circular shapes are not in the training distribution.}
    \label{fig:unet_circle_disturbance}
\end{figure}

For this analysis, a circle with a fixed radius of $0.1$ is randomly placed in the obstacle map. This modification is intended to examine how the network adapts its predictions when the obstacle map changes. In particular, we expect deviations in the predicted path when the newly added circle intersects with the path predicted based on the original obstacle configuration. Figure \ref{fig:unet_circle_disturbance} illustrates both the network’s predictions before and after adding the circle, alongside the path calculated by the RI-RRT* algorithm for the unmodified environment. We observe that the network successfully re-routes the path to avoid collision with the circle, producing reasonable trajectories even when the obstacle geometry includes shapes not present in the training set. This demonstrates its ability to handle deviating geometric configurations of the obstacle map and still predict reasonable trajectories.

\subsection{Path Quality}
While these visual comparisons highlight the network's adaptability, they offer limited information about the quality of the predicted paths in terms of Problem \ref{prob}. To quantitatively assess our approach's ability to produce accurate paths, we compared the path lengths of the reconstructed trajectories from both the neural network predictions and the RI-RRT* algorithm.
\begin{figure}[]
    \centering
    \includegraphics[width=\linewidth]{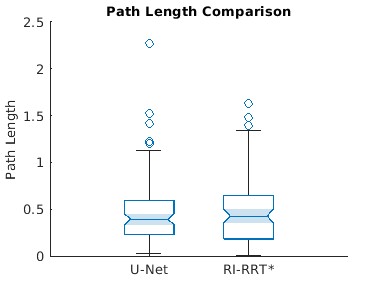}
    \caption{Distribution of resulting path lengths estimated by $100$ sample scenarios computed using the U-Net outputs and using RI-RRT*. In the plot, the medians are shown as the center lines in the shaded areas, and the 25\% and 75\% percentiles as the top and bottom lines of the box. Potential outliers are shown as circles outside the ``whiskers", reflecting the spread and skewness of the distributions.}
    \label{fig:path_length}
\end{figure}
Figure \ref{fig:path_length} shows the distribution of path lengths resulting from the RI-RRT* algorithm and the U-Net prediction. The path length, being part of the objective function in Problem \ref{prob}, serves as a natural metric for solution quality. We observe that the path lengths generated by both methods are comparable, except for some outliers in the neural network approach. The similarity in path lengths demonstrates that the network can produce solutions of comparable quality to those generated by the RI-RRT* algorithm.

\subsection{Computational Efficiency}
However, as stated above, the primary advantage of the neural network is not just accuracy but computational efficiency. 
% To quantify this claim we compare CPU times of both algorithms which is illustrated in Fig. % \ref{fig:comp_time}.
% \begin{figure}[]
%     \centering
%     \includegraphics[width=\linewidth]{images/CompTime.png}
%     \caption{Comparison of CPU runtime for different version of the algorithm.}
%     \label{fig:comp_time}
% \end{figure}
Construction of a feasible near-optimal path from the density computed by our network is
%The neural network computes approximate optimal paths in an average of 5 seconds, which is 
approximately $12$ to $15$ times faster than the traditional method.
%$60-75$ second runtime. 
However, our current implementation for 
path reconstruction from the image space (U-Net's output) has substantial room for acceleration. We currently use Dijkstra's algorithm on a fully connected graph, which could be enhanced by employing the $A^\star$ algorithm and restricting edge connections to nearby nodes. About $70\%$ of the computational time is consumed by graph construction, primarily due to inefficient collision checks on image representations. These checks are considerably more efficient when obstacles are represented as triangles, as in the traditional method. With these optimizations implemented, we anticipate that the neural network approach could potentially achieve at least a $50\times$ speed improvement over the traditional method.

\section{Conclusion\label{sec:conclusion}}
In this paper, we presented a novel approach to belief space planning using deep learning to predict optimal feasible paths under Gaussian uncertainty directly. Our method, utilizing a U-Net architecture, significantly reduces computational complexity while maintaining path quality compared to traditional sampling-based planners. We introduced an efficient image encoding scheme and a reconstruction algorithm to translate neural network outputs into feasible paths. 
We
%Numerical simulations 
demonstrated the efficacy of our approach in terms of computational efficiency, path quality, and generalizability. Future work will focus on extending the method to higher-dimensional spaces.

\section{Acknowledgements}
This work is partially supported by the Air Force Office of Scientific Research under Grant FA9550-20-1-0101 and the Army Research Office under Grant W911NF-23-2-0240.

\bibliography{main.bib}

\end{document}